\documentclass[letterpaper]{article} 
\usepackage{aaai24}  
\usepackage{times}  
\usepackage{helvet}  
\usepackage{courier}  
\usepackage[hyphens]{url}  
\usepackage{graphicx} 
\usepackage{natbib}  
\usepackage{caption} 
\usepackage[utf8]{inputenc}
\usepackage{algorithm}
\usepackage{algpseudocode}
\usepackage{lipsum}

\usepackage{newfloat}
\usepackage{listings}
\DeclareCaptionStyle{ruled}{labelfont=normalfont,labelsep=colon,strut=off} 
\floatstyle{ruled}
\newfloat{listing}{tb}{lst}{}
\floatname{listing}{Listing}
\newcommand{\sys}{TinyM$^2$Net-V3}
\title{TinyM$^2$Net-V3: Memory-Aware Compressed \underline{M}ulti\underline{m}odal Deep Neural Networks for Sustainable Edge Deployment}

\author {
    Hasib-Al Rashid\textsuperscript{\rm 1,\rm 2},
    Tinoosh Mohsenin\textsuperscript{\rm 2}
}
\affiliations {
    \textsuperscript{\rm 1}University of Maryland, Baltimore County\\
    \textsuperscript{\rm 2}Johns Hopkins University\\
    hrashid1@umbc.edu, tinoosh@jhu.edu
}

\usepackage{bibentry}

\begin{document}

\maketitle

\begin{abstract} 

The advancement of sophisticated artificial intelligence (AI) algorithms has led to a notable increase in energy usage and carbon dioxide emissions, intensifying concerns about climate change. This growing problem has brought the environmental sustainability of AI technologies to the forefront, especially as they expand across various sectors. In response to these challenges, there is an urgent need for the development of sustainable AI solutions. These solutions must focus on energy-efficient embedded systems that are capable of handling diverse data types even in environments with limited resources, thereby ensuring both technological progress and environmental responsibility. Integrating complementary multimodal data into tiny machine learning models for edge devices is challenging due to increased complexity, latency, and power consumption. This work introduces \emph{\sys{}}, a system that processes different modalities of complementary data, designs deep neural network (DNN) models, and employs model compression techniques including knowledge distillation and low bit-width quantization with memory-aware considerations to fit models within lower memory hierarchy levels, reducing latency and enhancing power efficiency on resource-constrained devices. We evaluated \emph{\sys{}} in two multimodal case studies: COVID-19 detection using cough, speech, and breathing audios, and pose classification from depth and thermal images. With tiny inference models (6 KB and 58 KB), we achieved 92.95\% and 90.7\% accuracies, respectively. Our tiny machine learning models, deployed on resource limited hardware, demonstrated low latencies within milliseconds and very high power efficiency.
\end{abstract}

\maketitle

\vspace{-3ex}
\section{Introduction}
As artificial intelligence (AI) continues to advance, energy-intensive AI algorithms are increasingly gaining traction. The quest for greater accuracy in solving complex, large-scale issues has prompted the use of deeper and more complex AI models. These models, while effective, come with a significant environmental cost. They require extensive computational power, leading to increased energy use and, consequently, higher carbon dioxide emissions, which are a major contributing factor to climate change. To put this into perspective, a study highlighted that training a single advanced natural language processing (NLP) model using deep learning techniques can produce as much as 626,000 pounds of carbon dioxide \cite{kraus2023enhancing}. This figure underscores the environmental impact of these advanced AI technologies, emphasizing the need for more sustainable approaches in the field of artificial intelligence. The growing concern over increasing carbon emissions and global waste has heightened the urgency for sustainable AI solutions.

Moreover, the United Nations has established the 2030 Agenda for Sustainable Development, a comprehensive framework focused on fostering peace and prosperity, which is anchored by 17 Sustainable Development Goals (SDGs) \cite{UN}. These goals serve as a universal call to action for all countries to strive for a future that balances environmental, economic, and social sustainability. In response to this, Edge Machine Learning (EdgeML) and Tiny Machine Learning (TinyML) have risen as sustainable alternatives, enabling the execution of machine learning models on smaller, lower-powered devices like mobile phones, wearables, and Internet of Things (IoT) devices \cite{prakash2023tinyml}. EdgeML and TinyML can significantly contribute to achieving various SDGs, especially those related to environmental sustainability.

However, designing accurate and efficient models for these devices is challenging due to their limited computing and memory resources. Model compression techniques, including pruning, quantization, and knowledge distillation, have been widely used to address these challenges by reducing the size and computational complexity of the models. Moreover, most of these model compression techniques target unimodal models to be compressed for sustainable edge hardware deployment. Potentials of multimodal deep neural networks (M-DNN), processing multiple modalities of complementary data are thus ignored for tiny device deployment. M-DNNs, which combine information from multiple sources such as text, image, and audio, have shown great potential in various applications such as speech recognition, natural language processing, and autonomous driving. However, Implementing M-DNN models in resource-limited edgeML and tinyML applications is challenging due to the growing number of model parameters and computations.
These challenges further complicate the task of implementing efficient and sustainable M-DNN inference while maintaining low peak memory consumption.

\begin{figure}
\centerline{\includegraphics[width=0.42\textwidth]{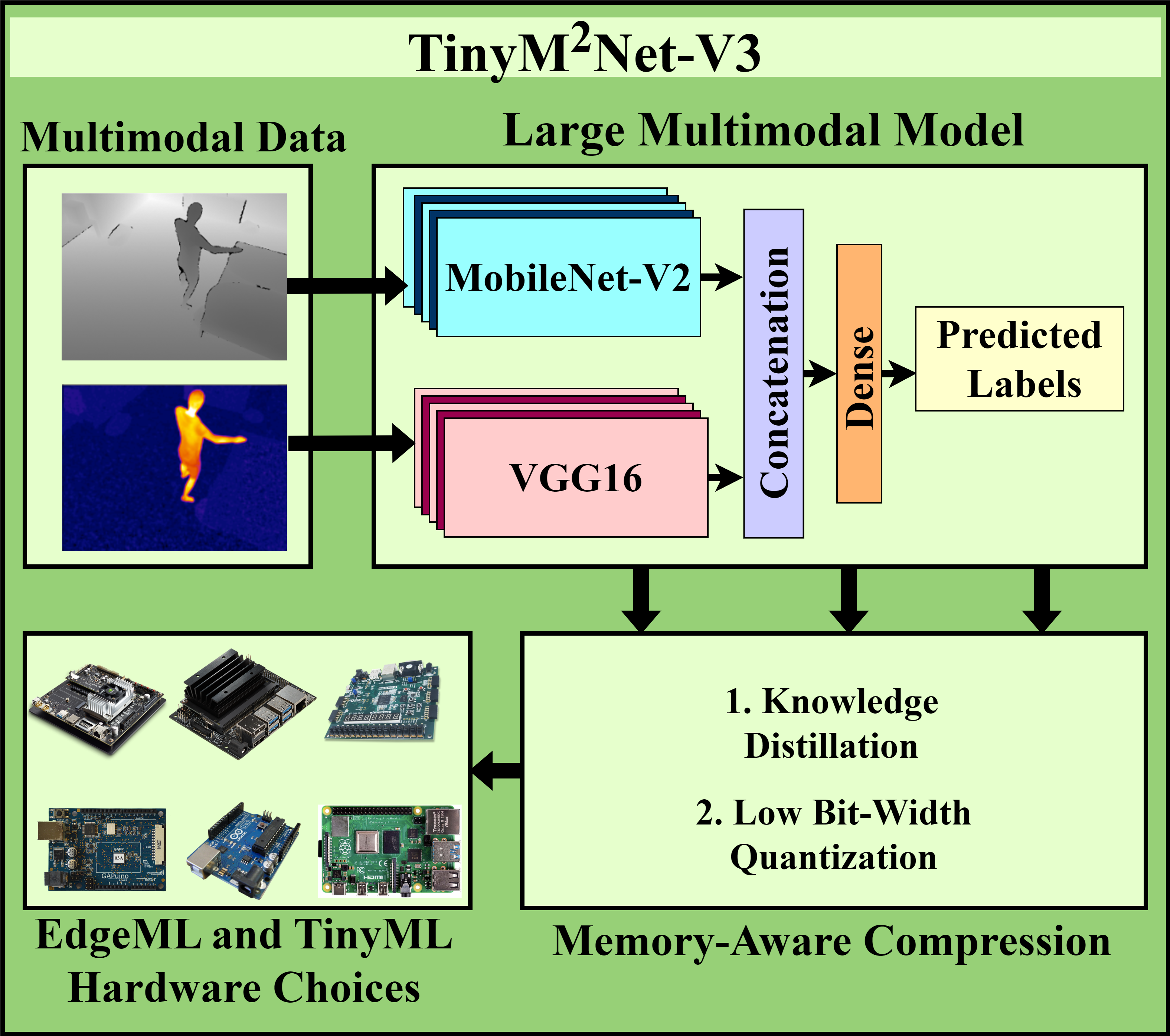}}
\caption{\small The high-level overview of the proposed \sys{}. \sys{} is capable of handling any number of data modalities, designing ML models for specific tasks, compressing the models using state-of-the-art compression techniques knowledge distillation and low bit-width quantization, and subsequently deploying them on resource-constrained tiny hardware.}
\label{highlevel}
\end{figure}

In this paper, we tackle the challenge of implementing M-DNN models on various resource-constrained hardware. To achieve energy-efficient M-DNN models on tiny processing hardware, we leverage the advantages of state-of-the-art compression techniques including knowledge distillation and low bit-width quantization. We propose a hardware aware, to be specific memory aware knowledge distillation and quantization technique that reduces the model size significantly while maintaining the model accuracy based on application needs. The high-level overview of the proposed \emph{\sys{}} is illustrated in Figure \ref{highlevel}. We assess \emph{\sys{}} in two multimodal case studies: COVID-19 detection with multimodal audios and pose classification using multimodal images. \emph{\sys{}} is subsequently implemented onto GAPuino and Raspberry Pi 4B to evaluate real-time performance on diverse resource-constrained hardware. The primary contributions of this paper are as follows:

\begin{itemize}
 \item We propose \emph{\sys{}}, an end-to-end hardware aware algorithms for multimodal neural networks suitable for sustainable edge deployment. \emph{\sys{}} introduces multimodal data (images and audios) to be adapted in sustainable ML models to improve the application specific accuracies while maintaining required performance metrics for sustainable edge deployment.
 \item We compress the M-DNN models with hardware aware knowledge distillation and uniform 8-bit quantization to reduce memory consumption and computational complexity for sustainable edge deployment.
 \item We evaluate the proposed \emph \sys{} for two different case-studies. \emph{Case-study 1} includes COVID-19 detection from multimodal audio recordings. \emph{Case-study 2} includes pose classification from multimodal depth and thermal images. 
\item We implemented our models on two resource-contrained hardware, GAPuino and Raspberry Pi 4B boards and explored their power efficiency with our multimodal models.
\end{itemize}

\section{Proposed \protect\sys{} System}

\begin{figure}
\centerline{\includegraphics[width=0.45\textwidth]{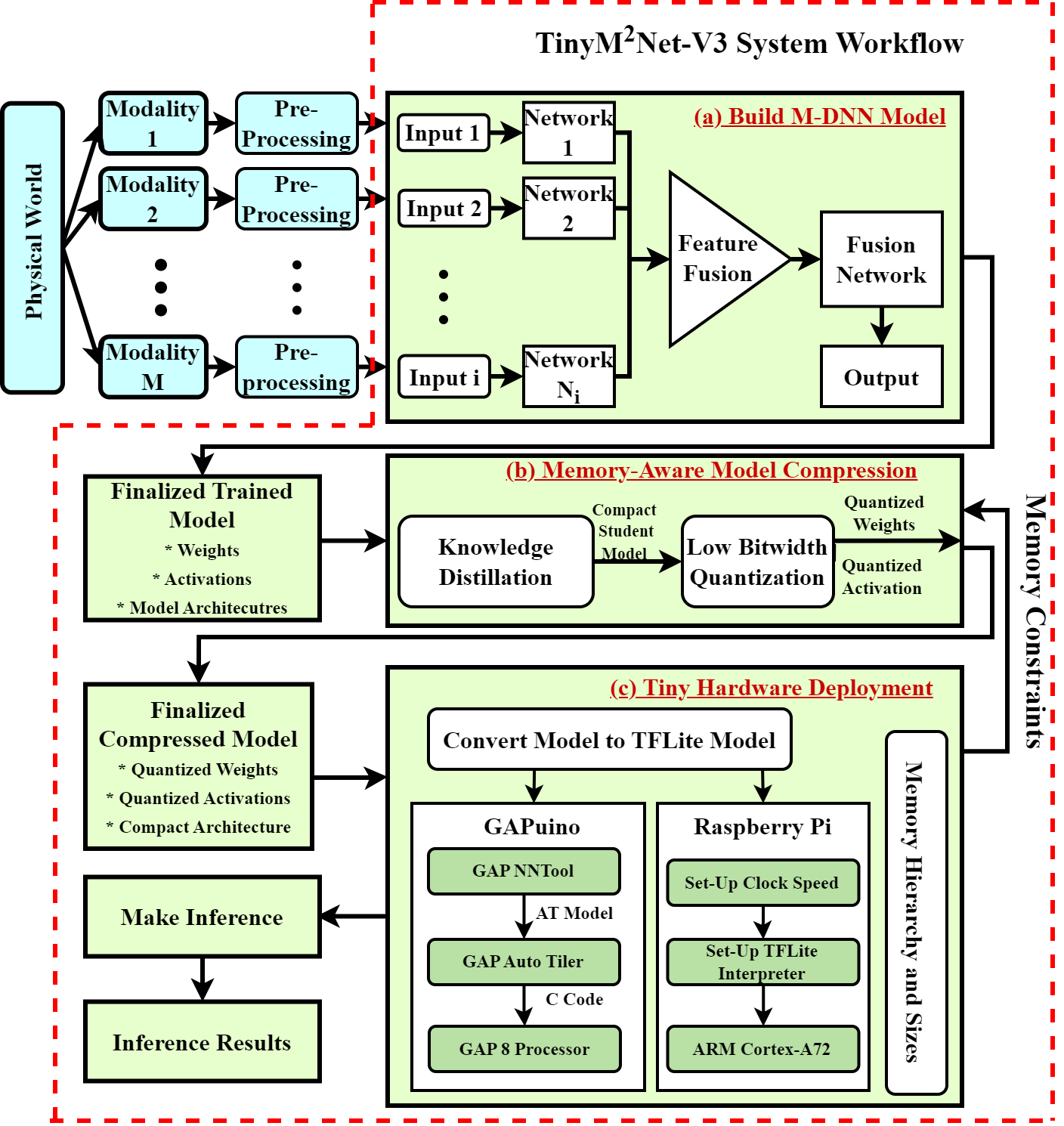}}
\caption{\small The flow diagram of proposed \sys{} system. We consider pre-processed multimodal inputs for our proposed \sys{}. Proposed \sys{} is the sequential combination of the steps shown in the diagram.}
\label{recipe}
\end{figure}

\subsection{Multimodal DNN Model Architecture Design}
Figure \ref{recipe} (a) illustrates the process of constructing Multimodal Deep Neural Network (M-DNN) models by incorporating various data modalities from the physical world and pre-processing them to suit the neural network formulation. The rationale behind utilizing multimodal data is to harness complementary information from each modality for a given learning task, ultimately leading to a more robust representation and superior results compared to employing a single modality. We formulated a multimodal learning problem in which multiple data modalities are exploited for classification tasks. In this work, we adopted the \emph{Intermediate fusion} technique to fuse multiple modalities, owing to its demonstrated superiority over other fusion methods \cite{stahlschmidt2022multimodal}.

In designing the multimodal model, we have designed unimodal models with each modalities. The hyperparameters (filter size, number of filters) for each unimodal network can be ascertained using neural architecture search (NAS) algorithms or can be assigned empirically. We chose the unimodal networks demonstrating the highest accuracies for each modality to concatenate. Upon obtaining features from each unimodal network, they are fused and forwarded to the fusion network. Ultimately, the classification output is represented as a probability distribution of the final fully connected layer using the Softmax activation function.

\subsection{Memory-Aware Model Compression}
Memory-aware model compression refers to a set of techniques that focus on reducing the memory footprint of deep learning models while maintaining their performance. This is particularly important when deploying models on resource-constrained devices, such as embedded systems, IoT devices, or tiny machine learning platforms, which often have limited memory capacity. Memory-aware model compression aims to optimize models to fit within the available memory resources of the target hardware platform while minimizing the impact on accuracy. We propose a memory-aware compression technique for M-DNNs including off-the-shelf knowledge distillation and quantization, that focuses on reducing the memory footprint of a M-DNN model while maintaining its accuracy. In this approach, a smaller student M-DNN model is trained to mimic the behavior of a larger, more complex and accurate teacher M-DNN model, with the primary goal of minimizing the memory requirements of the student model. Our goal is to reduce the student model down to the point where we can fit most of the model onto the on-chip memories (L1 and L2 memories) of the tiny processors. To this extent, we considered several factors in memory-aware model compression:

\begin{itemize}
 \item \textbf{Memory Hierarchy of the Deployment Hardware:} Memory-aware model compression in \sys{} takes into account the memory hierarchy of the target hardware platform, such as on-chip SRAM, off-chip DRAM, or Flash memory, to ensure that the compressed model can be effectively stored and executed within the available lower level of memory hierarchies for faster and more efficient deployment.
 \item \textbf{Model Compression Techniques:} To compress the large multimodal neural network models, \sys{} used off-the-shelf model compression  such as: Knowledge Distillation, Uniform 8-bit Quantization and Compact Network architecture design for inference.
\end{itemize}

\begin{figure}
\centerline{\includegraphics[width=0.45\textwidth]{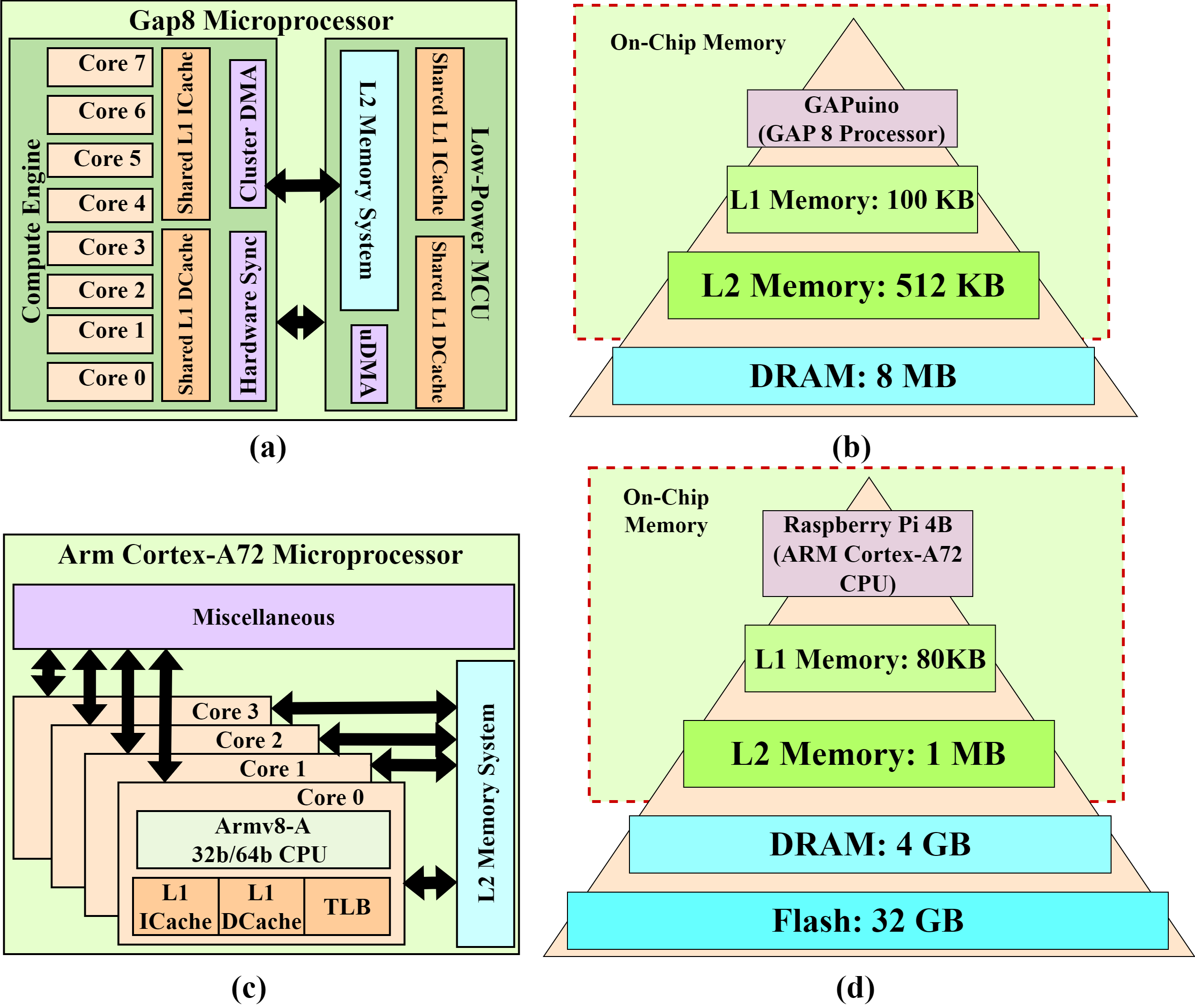}}
\caption{\small (a) Hardware Architecture for GAP8 microprocessor. (b) Memory Hierarchy of GAP8. GAP 8 microprocessor has L1 Memory of 100 KB (80 KB shared in compute engine + 20 KB for low power MCU.), l2 memory of 512 KB and 8MB of DRAM (c) Hardware Architecture for Arm Cortex-A72 microprocessor used in Raspberry Pi 4B. (d) Memory Hierarchy of ARM Cortex-A72 CPU, which has L1 Memory of 80 KB (48 KB Instruction Cache + 32 KB Data Cache), L2 memory of 1 MB, DRAM of 4 GB and external flash was 32 GB}
\label{hm}
\end{figure}

\begin{figure*}
\centerline{\includegraphics[width=\textwidth]{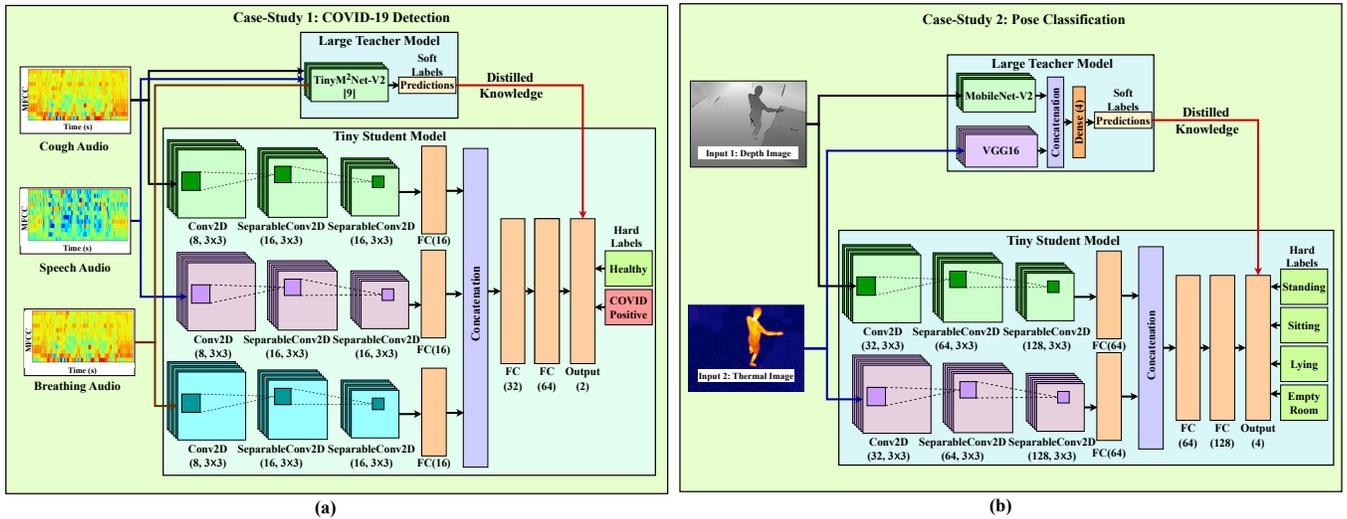}}
\caption{\small The model architecture of the proposed \emph{\sys{}} for (a) Case-study 1 and (b) Case-study 2. Here, Conv2D = 2 dimensional CNN, SeparableConv2D = 2 dimensional depthwise-separable CNN and FC = Fully Connected Layer.}
\label{models2}
\end{figure*}

Knowledge Distillation involves training a smaller student model to mimic the behavior of a larger, more accurate teacher model. The student model should have a smaller size than the teacher model, with fewer layers, parameters, or connections, to reduce memory requirements and enable deployment on memory-constrained devices. Different distillation techniques can be utilized to transfer knowledge from the teacher to the student model, such as soft targets, attention transfer, or feature map matching. We selected soft targets to transfer the teacher knowledge into student. This is achieved by using the soft targets generated by the larger model as training labels for the smaller model. The soft targets are obtained by applying a temperature scaling factor, $T$, to the output probabilities of the larger model, which smooths out the peaks and makes the distribution more spread out. More formally, let us denote the output probabilities of the larger model as $p_i$ and the soft targets as $q_i$. The soft targets are defined as follows:
\begin{equation}
q_i = \frac{\mathrm{exp}(z_i/T)}{\sum_j \mathrm{exp}(z_j/T)}
\end{equation}

where $z_i$ is the logit (unnormalized log-probability) output of the larger model for class $i$. The temperature scaling factor, $T$, controls the "softness" of the targets, with higher values of $T$ resulting in softer targets. The smaller model is then trained to minimize the Kullback-Leibler (KL) divergence between its output probabilities, $p_i'$, and the soft targets, $q_i$:

\begin{equation}
\mathcal{L} = \sum_i q_i \log\frac{q_i}{p_i'}
\end{equation}

where $p_i'$ is the output probability of the smaller model for class $i$. The KL divergence measures the difference between two probability distributions, and the loss function encourages the smaller model to learn a similar distribution to that of the larger model.

\par Designing compact and efficient network architectures can help create models with smaller memory footprints without sacrificing accuracy. We have used depthwise separable convolution layers in stead of regular convolution layers to further compress the student model without significant loss of accuracy.

 \par Reducing the bit-width of the model parameters and activations can significantly reduce the memory footprint of the model. Hardware-agnostic quantization might employ techniques such as uniform or mixed-precision quantization to minimize the memory footprint and computational complexity of the model. However, without considering the target hardware's specific capabilities and constraints, the quantized model may not be optimized for efficient execution on the target device, which could lead to suboptimal performance or incompatibilities. Hardware-aware quantization, on the other hand, can leverage ths knowledge to select the most suitable bit-width for the model's parameters and activations, ensuring better performance on the target device without sacrificing accuracy. Therefore, we adopted hardware aware quantization where we quantized our model to uniform 8-bits as both of our targeted hardware choices, Gap8 processor and Arm Cortex-A72 supports int8 data types for efficient computations. We quantized our models with Tensorflow Lite (tf-lite) post-training quantization adopting full integer quantization. This method quantizes both the weights and activations to 8-bit integers, resulting in a model that performs only integer arithmetic.

\subsection{Deployment on Resource-Constrained Hardware}
GAPuino development board was used in this work as main targeted sustainable edge deployment hardware, which is a nona-core 32-bit RISC-V ultra-low-power microprocessor for edge computing and IoT applications.

The GAP8 comprises autonomous peripherals, an ultra-low-power micro-controller, and a compute engine.
The GAP8 features a dedicated L1 cache for the MCU core, which includes 16 KB of data cache and 4 KB of instruction cache. Furthermore, the compute engine consists of eight additional cores sharing the same 64 KB data and 16 KB instruction caches, operating on separate voltage and frequency domains to optimize power consumption. In addition to these caches, the entire chip shares a 512 KB L2 cache, which is divided into four 128 KB cache banks. GAP8 hardware architecture and its memory hierarchy are shown in the figure \ref{hm} (a) and (b). GAP8 offers software libraries optimized for deep learning, image processing, data analysis, and encryption. With a real-time clock (RTC) for low-power standby modes, GAP8 is well-suited for energy-efficient, battery-powered edgeML and tinyML applications.

Figure \ref{recipe}(c) highlights the use of the GAPFlow toolchain in this work, consisting of NNTOOL and AutoTiler. NNTOOL adapts the DNN architecture, ensuring compatibility with AutoTiler and transforming weights for GAP8. AutoTiler algorithmically optimizes memory layout and generates GAP8-compatible C code. Despite automation, manual adjustments are occasionally needed for specific DNNs, such as modifying maximum stack sizes or adjusting heap space. By default, AutoTiler allocates the entire L1 and L2 memory, potentially causing heap overflows, data corruption, and stack issues. The GAP8's Real-Time Operating System (RTOS) further complicates matters by allocating heap memory before DNN initialization, reducing available space.

In this work, the Raspberry Pi 4B, featuring an Arm Cortex-A72 microprocessor, serves as a secondary sustainable edge deployment platform to compare the performance of the \sys{} system on an edge device. The ARMv8-A architecture-based 64-bit Arm Cortex-A72 microprocessor supports scalable multicore configurations and has a memory hierarchy consisting of separate L1 caches for each core and a shared L2 cache, enhancing performance and power efficiency. Figure \ref{hm} (c) and (d) illustrate the Cortex-A72 hardware architecture and memory hierarchy.

\section{\protect\sys{} Evaluation Results and Analysis}

\begin{figure*}
\centerline{\includegraphics[width=\textwidth]{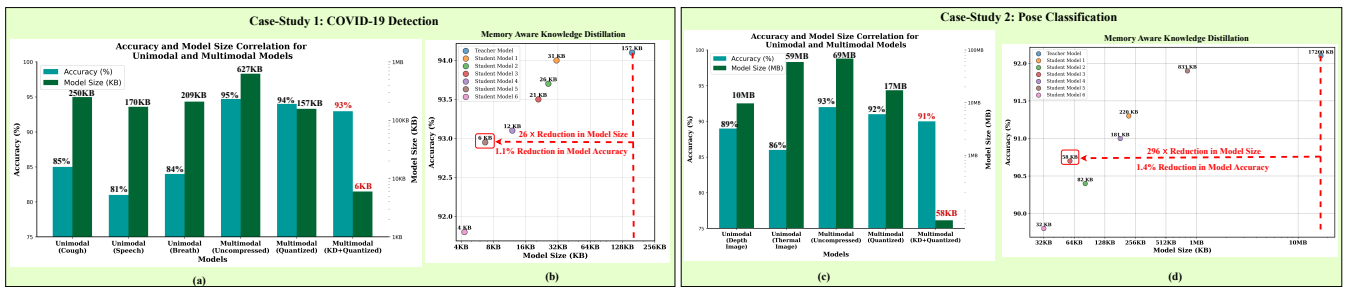}}
\caption{\small (a) \sys{} classification results for Case-Study 1 in terms of both unimodal and multimodal settings. The multimodal setting improved 7\% accuracy compared to unimodal (speech) classification setting. Model compression techniques reduce 1.6\% accuracy of the multimodal setting. (b) Experiments for memory-aware knowledge distillation. (a) \sys{} classification results for Case-Study 2 in terms of both unimodal and multimodal settings. The multimodal setting improved 6\% accuracy compared to unimodal (thermal) classification setting. Model compression techniques reduce 1.4\% accuracy of the multimodal setting. (d) Experiments for memory-aware knowledge distillation.}
\label{results}
\end{figure*}

\subsection{Evaluation Case-Study 1: COVID-19 Detection from Multimodal Audios}

For this case-study, we utilized the second DiCOVA challenge dataset \cite{sharma2021second}, a Coswara dataset subset \cite{sharma2020coswara}, comprising 929 participants with cough, breathing, and speech audio samples, including 172 COVID-positive cases. We classified COVID-19 test results as positive (`P') or negative (`N') and generated 6000 balanced, two-second MFCC spectrogram samples for \sys{} input. We adopted a 22.5 KHz sampling frequency, allocating 70\% data for training, 10\% for validation, and 20\% for evaluation.
\emph{\sys{}} processed three modalities using parallel CNN layers, extracting and fusing features for binary classification. We based the teacher model on TinyM$^2$Net-V2 \cite{rashid2023tinym2netv2} and designed the student model from scratch, as shown in figure \ref{models2}(a). We trained the model for 200 epochs using categorical cross-entropy loss and the Adam optimizer, assessing performance with accuracy metrics.

Figure \ref{results}(a) presents Case-Study 1 evaluation results. Single modality data yielded lower binary classification accuracies compared to the uncompressed multimodal model. Incorporating multimodal audio data increased model accuracy to 94.7\%, with the 8-bit quantized model achieving 94.1\% accuracy. We aimed to design a student model compressible to a few KB, fitting L1 and L2 caches of our deployment hardware. We experimented with filter sizes, dense layer neuron numbers, and replacing 2nd and 3rd CNN layers with depthwise separable counterparts. Student Model 5 reached 93\% accuracy with a 6 KB model size, as shown in figure \ref{results}(b). We selected this final student model for inference, achieving a 26$\times$ reduction in model size compared to its teacher model.

\subsection{Evaluation Case-Study 2: Pose Classification from Multimodal Depth Images and Thermal Images}
The authors in \cite{9191284} introduced the synthetic multi-modal "Sdt" dataset for pose classification tasks, which serves as a benchmark for evaluating algorithms using depth and thermal image modalities. The dataset comprises 40,000 images per modality, depicting individuals in standing, sitting, and lying poses, as well as empty rooms. Images are resized to 64x64 to reduce hardware memory demands, and the data is split into 70\% for training, 10\% for validation, and 20\% for testing. \emph{\sys{}} utilizes parallel CNN layers to process depth and thermal modalities, employing MobileNet-V2 and VGG16, respectively, to extract and fuse features for multi-class classification. The teacher model, illustrated in figure \ref{models2} (b), adopts pre-trained ImageNet weights and is trained for 200 epochs using categorical cross-entropy loss and the Adam optimizer. The teacher model achieves 92.6\% accuracy in pose classification.

Figure \ref{results}(c) presents \sys{} evaluation results for Case-Study 2. Single modality data yields lower pose classification accuracies compared to the uncompressed multimodal model. Incorporating multimodal image data improves accuracy to 92.3\%, while the 8-bit quantized multimodal model achieves 91.7\% accuracy. We then experimented to design the student model from scratch incorporating memory aware knowledge distillation. Our target was to compress the student model down to some KB so that we could fit the model on the L1 and L2 caches of the hardware we used for deployment. To this end we experimented with different filter sizes, number of neurons in dense layers and also replacing the 2nd and 3rd CNN layers to their depthwise separable counterparts so that we could achieve ultimate compression for the student model. Student Model 5 achieves around 90.9\% of accuracy with only 58 KB of model size. The experimental results are shown in the figure \ref{results}(d). We selected this as our final student model for inference which achieves 296$\times$ reduction in model size from its teacher model at the cost of 1.4\% reduction in model accuracy.

\section{\protect\sys{} Hardware Implementation Results and Analysis}
\begin{table}[]
\caption{\small Resource utilization data of \sys{} implemented on GAP8 Processor}
\label{res}
\centering
\scalebox{0.85}{
\begin{tabular}{|c|c|c|c|}
\hline
\textbf{Resources}                                                               & \textbf{L1 Memory} & \textbf{L2 Memory} & \textbf{DRAM} \\ \hline
\textbf{\begin{tabular}[c]{@{}c@{}}Available for Use\\ (KB)\end{tabular}}        & 52.7               & 400                & 8000          \\ \hline
\textbf{\begin{tabular}[c]{@{}c@{}}Case-Study 1\\ Utilization (KB)\end{tabular}} & 52.4 (99\%)        & 40 (10\%)          & 0             \\ \hline
\textbf{\begin{tabular}[c]{@{}c@{}}Case-Study 2\\ Utilization (KB)\end{tabular}} & 47.5 (90\%)        & 178 (44\%)         & 0             \\ \hline
\end{tabular}}
\vspace{-2ex}
\end{table}


\begin{figure}
\centerline{\includegraphics[width=0.45\textwidth]{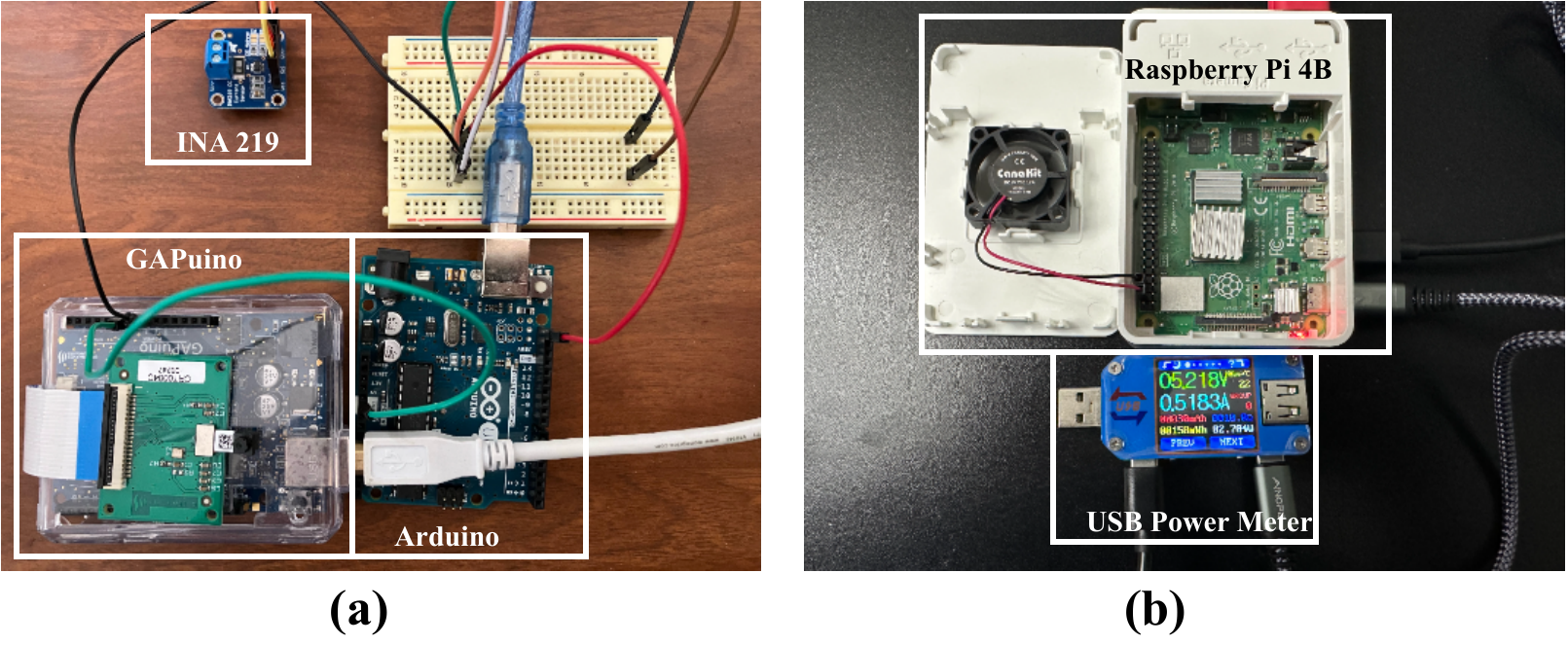}}
\caption{\small (a) GAPuino board power measurement setup. INA219 and Arduino measure the GAP8 power consumption. (b) Raspberry Pi 4B power measurement setup. USB power measurement device was used for raspberry Pi 4B.}
\label{power}
\end{figure}

\begin{table*}[]
\centering
\caption{\small Implementation Results of the proposed \sys{} and Comparisons with Previous Works.}
\label{comp_table}
\scalebox{0.80}{
\begin{tabular}{|c|cccc|cc|cc|}
\hline
\textbf{Architectures} &
  \multicolumn{4}{c|}{\textbf{This Work}} &
  \multicolumn{2}{c|}{\cite{rashid2022tinym2net}} &
  \multicolumn{2}{c|}{\cite{rashid2023tinym2netv2}} \\ \hline
\textbf{Application} &
  \multicolumn{2}{c|}{\begin{tabular}[c]{@{}c@{}}COVID-19\\ Detection\end{tabular}} &
  \multicolumn{2}{c|}{\begin{tabular}[c]{@{}c@{}}Pose\\ Classification\end{tabular}} &
  \multicolumn{1}{c|}{\begin{tabular}[c]{@{}c@{}}COVID-19\\ Detection\end{tabular}} &
  \begin{tabular}[c]{@{}c@{}}Object\\ Classification\end{tabular} &
  \multicolumn{1}{c|}{\begin{tabular}[c]{@{}c@{}}Vehicle\\ Clasifications\end{tabular}} &
  \begin{tabular}[c]{@{}c@{}}COVID-19\\ Detection\end{tabular} \\ \hline
\textbf{Modality Used} &
  \multicolumn{2}{c|}{\begin{tabular}[c]{@{}c@{}}Audio+Audio\\ +Audio\end{tabular}} &
  \multicolumn{2}{c|}{Image+Image} &
  \multicolumn{1}{c|}{Audio+Audio} &
  Image+Audio &
  \multicolumn{1}{c|}{Image+Audio} &
  \begin{tabular}[c]{@{}c@{}}Audio+Audio\\ +Audio\end{tabular} \\ \hline
\textbf{Operations (GOP)} &
  \multicolumn{2}{c|}{0.55} &
  \multicolumn{2}{c|}{2.38} &
  \multicolumn{1}{c|}{-} &
  - &
  \multicolumn{1}{c|}{0.42} &
  0.01 \\ \hline
\textbf{Edge Devices} &
  \multicolumn{1}{c|}{\begin{tabular}[c]{@{}c@{}}Raspberry Pi\\ 4B\end{tabular}} &
  \multicolumn{1}{c|}{GAPuino} &
  \multicolumn{1}{c|}{\begin{tabular}[c]{@{}c@{}}Raspberry Pi\\ 4B\end{tabular}} &
  GAPuino &
  \multicolumn{2}{c|}{\begin{tabular}[c]{@{}c@{}}Raspberry Pi\\ 4B\end{tabular}} &
  \multicolumn{2}{c|}{\begin{tabular}[c]{@{}c@{}}Raspberry Pi\\ 4B\end{tabular}} \\ \hline
\textbf{Frequency (MHz)} &
  \multicolumn{1}{c|}{1500} &
  \multicolumn{1}{c|}{175} &
  \multicolumn{1}{c|}{1500} &
  175 &
  \multicolumn{2}{c|}{1500} &
  \multicolumn{2}{c|}{1500} \\ \hline
\textbf{Latency (ms)} &
  \multicolumn{1}{c|}{\textbf{2.75}} &
  \multicolumn{1}{c|}{\textbf{4.47}} &
  \multicolumn{1}{c|}{\textbf{9.95}} &
  \textbf{49.20} &
  \multicolumn{1}{c|}{1200} &
  798 &
  \multicolumn{1}{c|}{1240} &
  980 \\ \hline
\textbf{Power (mW)} &
  \multicolumn{1}{c|}{\textbf{620}} &
  \multicolumn{1}{c|}{\textbf{215.7}} &
  \multicolumn{1}{c|}{\textbf{787}} &
  \textbf{307.6} &
  \multicolumn{1}{c|}{1700} &
  959 &
  \multicolumn{1}{c|}{1567} &
  994 \\ \hline
\textbf{Energy (mJ)} &
  \multicolumn{1}{c|}{1.70} &
  \multicolumn{1}{c|}{0.95} &
  \multicolumn{1}{c|}{7.83} &
  15.13 &
  \multicolumn{1}{c|}{2040} &
  765.28 &
  \multicolumn{1}{c|}{1800} &
  974.12 \\ \hline
\textbf{Performance (GOP/s)} &
  \multicolumn{1}{c|}{200} &
  \multicolumn{1}{c|}{124.43} &
  \multicolumn{1}{c|}{239.20} &
  48.37 &
  \multicolumn{1}{c|}{-} &
  - &
  \multicolumn{1}{c|}{0.33} &
  0.01 \\ \hline
\textbf{Power-Efficiency (GOP/s/W)} &
  \multicolumn{1}{c|}{\textbf{322.5}} &
  \multicolumn{1}{c|}{\textbf{576.8}} &
  \multicolumn{1}{c|}{\textbf{303.93}} &
  \textbf{157.26} &
  \multicolumn{1}{c|}{-} &
  - &
  \multicolumn{1}{c|}{0.22} &
  0.01 \\ \hline
\end{tabular}
}
\end{table*}

To evaluate the \sys{} approach, we deployed the trained models on the GAP8 processor. Table \ref {res} reports resource utilization of \sys{} for 2 case-studies implemented separately on GAP8 processor. Case-Study 1 uses 52.4 KB of L1 memory and 40 KB of L2 memory which is only 10\% of the available L2 memory. The inference model does not require off-chip DRAM to store its weights and activations which ensures the minimum latency. Similarly, Case-Study 2 uses 47.5 KB of L1 memory and 178 KB of L2 memory which is 44\% of the available L2 memory. This inference model as well does not require off-chip DRAM to store its weights and activations which ensures the minimum latency. 


Figure \ref{power} (a) displays the power measurement setup used in this work for GAPuino board, using INA 219 sensor and Arduino board and the figure \ref{power} (b) shows Raspberry Pi 4B power measurement setup where USB power measurement device was used.

Table \ref{comp_table} reports latency and power consumption of the \sys{} implemented on GAPuino and Raspberry Pi 4B Boards. Raspberry PI 4B ran on 1500 MHz clock frequency while GAP8 clock frequency was 175 MHZ. For both the case-studies, GAP8 implementations were running slower but consumed less power compared to Raspberry Pi 4B. Our Raspberry Pi implementation has power efficiency of 322.5 GOP/s/W and 303.93 GOP/s/W respectively for case-study 1 and 2. Our GAPuino implementation has power efficiency of 576.8 GOP/s/W and 157.26 GOP/s/W respectively for case-study 1 and 2. The Raspberry Pi might exhibit better power efficiency for case-study 2 as it has larger size of activation memory which caused delays in inference latency on GAP8 processor. The overall power-efficiency depends on the combination of hardware, software, and the model's characteristics. We have also compared our both the implementations with previous multimodal models deployed on resource-constrained hardware devices. As our work targets hardware-aware model compression, both of our case-studies outperforms previous implementations with hardware-agnostic compressed models.

\vspace{-2ex}
\section{Conclusion}
\vspace{-1.0ex}
While advanced AI algorithms are technologically advanced, they often fall short in sustainability, primarily due to their high energy consumption and the need for extensive computational resources. In an effort to address these concerns, this paper introduces \emph{\sys{}}, a novel system designed to process diverse modalities of complementary data. \emph{\sys{}} focuses on designing deep neural network (DNN) models that are more efficient in terms of size and power consumption. By employing model compression techniques such as knowledge distillation and low bit-width quantization, and integrating hardware-aware design principles, the system efficiently compresses these models. This approach allows the models to fit within the lower levels of the memory hierarchy, significantly reducing latency and enhancing power efficiency. This makes \emph{\sys{}} particularly suitable for deployment on resource-constrained tiny devices, offering a sustainable alternative to traditional, resource-intensive AI models. To assess the effectiveness of \emph{\sys{}}, we evaluated with two case studies involving multimodal analysis: detecting COVID-19 from multimodal cough, speech and breathing audios and detecting poses from multimodal depth and thermal images. Our results showed that, despite the utilization of tiny inference models, we were able to attain an accuracy of 92.95\% for the COVID-19 detection task using an inference model size of merely 6 KB and 90.7\% accuracy for the pose detection task using an inference model size of only 58 KB. Our tinyML models were deployed on two sustainable edge hardware platforms, namely Raspberry Pi and GAPuino development boards, attaining latencies within the range of a few milliseconds and power consumption in the milliwatt range.

\bibliography{eehpc.bib, micro_tinyml}

\begin{thebibliography}{9}
\providecommand{\natexlab}[1]{#1}

\bibitem[{Kraus et~al.(2023)Kraus, Bingler, Leippold, Schimanski, Senni, Stammbach, Vaghefi, and Webersinke}]{kraus2023enhancing}
Kraus, M.; Bingler, J.~A.; Leippold, M.; Schimanski, T.; Senni, C.~C.; Stammbach, D.; Vaghefi, S.~A.; and Webersinke, N. 2023.
\newblock Enhancing Large Language Models with Climate Resources.
\newblock \emph{arXiv preprint arXiv:2304.00116}.

\bibitem[{Prakash et~al.(2023)Prakash, Stewart, Banbury, Mazumder, Warden, Plancher, and Reddi}]{prakash2023tinyml}
Prakash, S.; Stewart, M.; Banbury, C.; Mazumder, M.; Warden, P.; Plancher, B.; and Reddi, V.~J. 2023.
\newblock Is TinyML Sustainable? Assessing the Environmental Impacts of Machine Learning on Microcontrollers.
\newblock \emph{arXiv preprint arXiv:2301.11899}.

\bibitem[{Pramerdorfer, Strohmayer, and Kampel(2020)}]{9191284}
Pramerdorfer, C.; Strohmayer, J.; and Kampel, M. 2020.
\newblock Sdt: A Synthetic Multi-Modal Dataset For Person Detection And Pose Classification.
\newblock In \emph{2020 IEEE International Conference on Image Processing (ICIP)}, 1611--1615.

\bibitem[{Rashid et~al.(2022)Rashid, Ovi, Busart, and Mohsenin}]{rashid2022tinym2net}
Rashid, H.-A.; Ovi, P.~R.; Busart, A., Carl~Gangopadhyay; and Mohsenin, T. 2022.
\newblock TinyM2Net: A Flexible System Algorithm Co-designed Multimodal Learning Framework for Tiny Devices.
\newblock \emph{ArXiv}.

\bibitem[{Rashid et~al.(2023)}]{rashid2023tinym2netv2}
Rashid, H.-A.; et~al. 2023.
\newblock TinyM$^2$Net-V2: A Compact Low Power Software Hardware Architecture for \underline{M}ulti\underline{m}odal Deep Neural Networks.
\newblock \emph{ACM Transactions on Embedded Computing Systems}.

\bibitem[{Sharma et~al.(2020)}]{sharma2020coswara}
Sharma, N.; et~al. 2020.
\newblock Coswara--A Database of Breathing, Cough, and Voice Sounds for COVID-19 Diagnosis.

\bibitem[{Sharma et~al.(2021)Sharma, Chetupalli, Bhattacharya, Dutta, Mote, and Ganapathy}]{sharma2021second}
Sharma, N.~K.; Chetupalli, S.~R.; Bhattacharya, D.; Dutta, D.; Mote, P.; and Ganapathy, S. 2021.
\newblock The Second DiCOVA Challenge: Dataset and performance analysis for COVID-19 diagnosis using acoustics.
\newblock \emph{arXiv preprint arXiv:2110.01177}.

\bibitem[{Stahlschmidt, Ulfenborg, and Synnergren(2022)}]{stahlschmidt2022multimodal}
Stahlschmidt, S.~R.; Ulfenborg, B.; and Synnergren, J. 2022.
\newblock Multimodal deep learning for biomedical data fusion: a review.
\newblock \emph{Briefings in Bioinformatics}, 23(2): bbab569.

\bibitem[{{United Nations}(2021)}]{UN}
{United Nations}. 2021.
\newblock The Sustainable Development Goals Report.
\newblock \url{https://sdgs.un.org/goals}.
\newblock Accessed: 2023-11-23.

\end{thebibliography}

\end{document}